\documentclass[11pt,a4paper]{article}
\usepackage[hyperref]{eacl2021}
\usepackage{times}
\usepackage{latexsym}
\usepackage{comment}
\usepackage{url}
\usepackage{amsmath}
\usepackage{CJKutf8}
\usepackage[boxed]{algorithm}
\usepackage{caption}
\usepackage{algpseudocode}

\aclfinalcopy % Uncomment this line for the final submission

{\begin{itemize}[topsep=0pt, partopsep=0pt] 
	\setlength{\itemsep}{0pt}%
	\setlength{\parskip}{0pt}%
}%
{\end{itemize}}

\usepackage{booktabs}
\usepackage{subcaption}
\usepackage{xspace}
\usepackage{amstext}
\usepackage{booktabs}
\usepackage{graphicx}
\usepackage{tabulary}
\usepackage{graphicx}
\usepackage{enumitem}
\usepackage{color}
\usepackage{lipsum}
\usepackage{amsfonts}

\usepackage{relsize}
\usepackage{bbm}
\usepackage{times}
\usepackage{epsfig}
\usepackage{amssymb}
\usepackage{makecell}
\usepackage{multirow}
\usepackage{bm}
\usepackage{placeins}

\title{Named Entity Recognition in the Style of Object Detection}

\author{Bing Li  \\
	 Microsoft \\
	libi@microsoft.com
}

\begin{document}
\maketitle

%%%%%%%%% ABSTRACT
\begin{abstract}

In this work, we propose a two-stage method for named entity recognition (NER), especially for nested NER. We borrowed the idea from the two-stage Object Detection in computer vision and the way how they construct the loss function. First, a region proposal network generates region candidates and then a second-stage model discriminates and classifies the entity and makes the final prediction. We also designed a special loss function for the second-stage training that predicts the entityness and entity type at the same time. The model is built on top of pretrained BERT encoders, and we tried both BERT base and BERT large models. For experiments, we first applied it to flat NER tasks such as CoNLL2003 and OntoNotes 5.0 and got comparable results with traditional NER models using sequence labeling methodology. We then tested the model on the nested named entity recognition task ACE2005 and Genia, and got F1 score of 85.6$\%$ and 76.8$\%$ respectively. In terms of the second-stage training, we found that adding extra randomly selected regions plays an important role in improving the precision. We also did error profiling to better evaluate the performance of the model in different circumstances for potential improvements in the future.

\end{abstract}

%%%%%%%%% BODY TEXT
\section{Introduction}

Named entity recognition (NER) is commonly dealt with as a sequence labeling job and has been one of the most successfully tackled NLP tasks. Inspired by the similarity between NER and object detection in computer vision and the success of two-stage object detection methods, it's natural to ask the question, can we borrow some ideas from there and simply copy the success. This work aims at pushing further our understanding in NER by applying a two-stage object-detection-like approach and evaluating its effects in precision, recall and different types of errors in variant circumstances. 

As a prototype for many other following object detection models, Faster R-CNN~\cite{ren2015faster} is a good example to illustrate how the two stages work. In the first stage, a region proposal network is responsible for generating candidate regions of interest from the input image and narrowing down the search space of locations for a more complex and comprehensive investigation process. The second stage work is then done by the regression and classification models, which look at finer feature maps, further adjust the size and position of the bounding box and tell whether there is an object and what type it belongs to. Admittedly a two-stage pipeline is often blamed for error propagation, however in the case of object detection it apparently brings much more than that.

We can easily see the analogy here in NER. To do NER in a similar way, first we find candidate ranges of tokens that are more likely to be entities. Second, we scrutinize into each of them, make a judgment if this is a true entity or not, and then do entity classification and also regression if necessary. Even though the search space in 1-D NER problem is significantly smaller than in a 2-D image, the benefits of a two-stage approach are still obvious. Firstly, with the better division of labor, the two components can focus on their specialized tasks which are quite different. More specifically, the region proposal part takes care of the context information and entity signals regardless of entity type, while the second part covers more on entity integrity and type characteristics. Secondly, since the region prediction has been given by the first stage, the second part of the model can take the global information of all the tokens in the region, instead of looking separately as in sequence labeling. Although a single token vector can also encode context information as in BERT~\cite{devlin2018bert} and other similar LMs~\cite{elmo2018, Yang2019XLNet, Yinhan2019Roberta, Vaswani2017transformer, radford2019GPT2}, having a global view of all relevant tokens is definitely an advantage and may provide more hints for the final decision. Thirdly, the model gets better interpretability, since it separates the concepts of entityness (how likely this mention is a qualified named entity) and entity classification, you get an interpretable probability for each entity prediction. Finally, another benefit of this method is that each entity prediction is independent, which makes it possible to be used in a nested entity recognition problem that can not be easily handled by sequence labeling approach, which can only predict one label for each token.

This paper is structured as follows. In Section \ref{section:model} we explain the model architecture, including two stages, region proposal and entity classification. Section \ref{section:related} reviews the past related works in NER, especially region based NER that are most similar to our work. Section \ref{section:experiment} explains the training process and evaluation results in detail on flat NER tasks. Section ~\ref{section:nestedexp} shows the training and evaluation on nested NER tasks. In Section \ref{section:ablation} we evaluate the importance of different parts of the model by ablation. And finally in Section ~\ref{section:error} we show the error analysis of the new method and compared it with the traditional sequence labeling methodology.

\section{Model Design}
\label{section:model}
We propose a two-stage method for NER. In the first stage, an entity region proposal network predicts highly suspected entity regions. In the second stage, another model takes the output candidates, extracts more features and then makes a prediction whether this candidate is a qualified entity or not and what type should be attached to this entity. Precision and recall would be evaluated end-to-end, in the same manner as traditional NER models.

\begin{figure}
	\centering
	\includegraphics[width=1\linewidth]{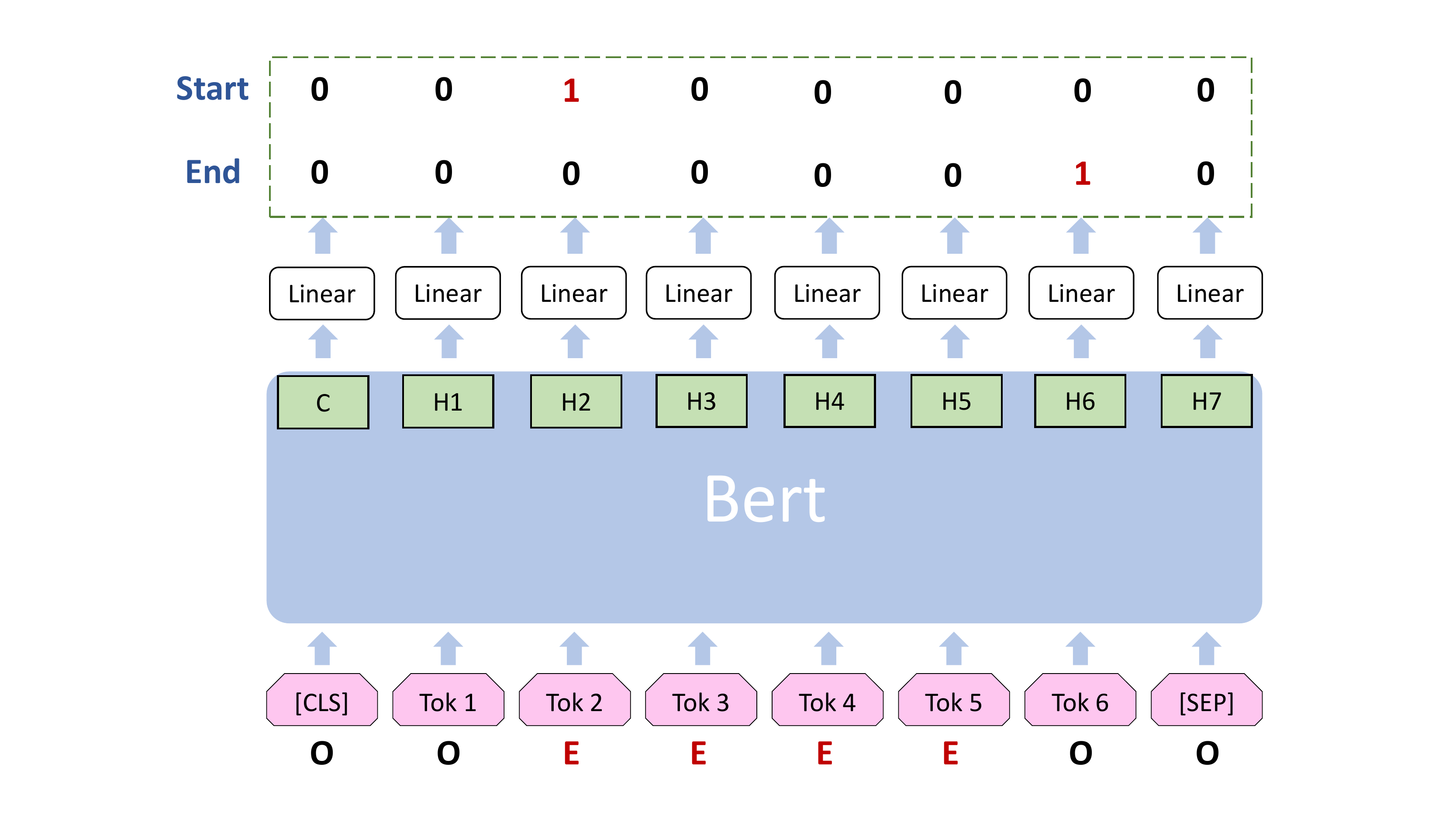}
	\caption{\small{Region proposal network is made by adding a linear layer on top of the BERT output. The start prediction is independent from the end prediction, both of which use cross entropy loss over two classes. For the end position, we assign it to the first token of the word right immediate after the entity. We only predict for the first token of every word if the tokenizer outputs WordPieces or other subword tokens, and all other tokens will be masked (the mask is omitted from the diagram).}}
	\label{fig:proposalNetwork}
\end{figure}

\subsection{Entity Region Proposal}
For the first stage, we used a simple model similar to sequence labeling ones. But instead of predicting IOB-format entity labels~\cite{ramshaw-marcus-1995-text}, we predict $\langle$Start$\rangle$ and $\langle$End$\rangle$ labels. We appended another linear layer on top of the hidden state outputs from the BERT model~\cite{devlin2018bert}, to predict the probability for a token to be $\langle$Start$\rangle$ and $\langle$End$\rangle$, we only consider the starting token in each word and mask out all trailing tokens within that word. And we used cross entropy over two classes for the prediction. The model structure is demonstrated in Figure \ref{fig:proposalNetwork}. The goal of the first-stage model is high recall and acceptable precision, so we could tilt the weights a little bit towards the positive labels to favor better recall numbers. After extracting the start and end tokens, we then select pairs to form a complete region, with the simple rule that the length of the entity cannot exceed some limit. Admittedly, there is an apparent disadvantage here that is we discard longer candidates without even a try. But in fact, those longer ones only compose quite a small portion and are usually very difficult to get right anyway. In practice we have used 6 and 12 as the length limit for different datasets and can easily cover 98$\%$-99$\%$ of ground-truth entities. 

\subsection{Entity Discrimination and Classification}

\begin{figure*}[!h]
	\centering
	\includegraphics[ width=0.92\linewidth ]{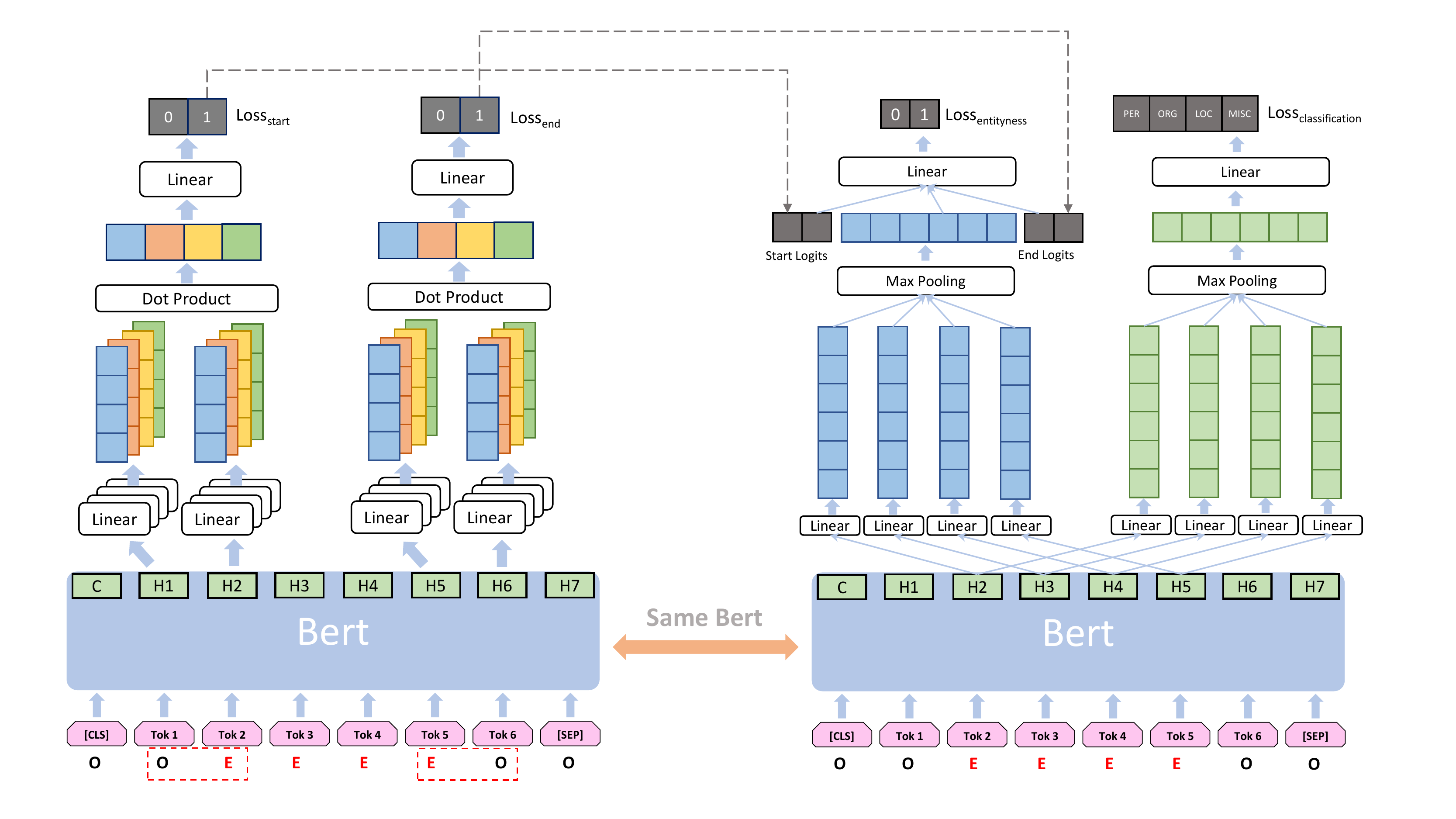}
	\caption{\small{The second-stage model is responsible for re-examining the entity region proposals generated by the first stage. The model trains with 4 tasks simultaneously. The entityness part and type classification part takes global view of all tokens within the range, while the two boundary losses only zoom into the two tokens across the boundary to make a double check of the exact range.}
	}
	\label{fig:mainModel}
\end{figure*}

The second stage is also using BERT~\cite{devlin2018bert} as the encoding model. The second stage has two main tasks, discrimination and classification, which defines the two major components in our loss function, the entityness loss and the type classification loss. Contrary to a typical object detection model, we don't do regression for the bounding box. The model will only tell if this range is correct or not and if not, discard it directly. One reason is that we want to keep it as simple as possible and another reason is that the problem is much easier than the 2-D object detection, and our proposals are usually accurate enough. Compared to the sequence labeling method, our model focuses more on aggregate features across all tokens spanning the entity, which can be seen from the max pooling layer in both the entityness and classification components. To make the model more sensitive to boundary errors, especially when coming across long entities, we added another two losses, the start loss and the end loss, which predict if the boundaries are correct, so that the prediction won't be dominated by the bulk part but also pay attention to the boundary. These start/end logits are also concatenated with the entityness feature to predict the entityness score. The total loss function can be written as below:

\scriptsize
\begin{align}
\textbf{L} &= \alpha \left(\textbf{L}_{\text{start}} + \textbf{L}_{\text{end}}\right) + \beta \textbf{L}_{\text{entityness}} + \textbf{L}_{\text{type}}
\\
\textbf{L}_{\text{s}} &= -\frac{1}{N}\sum_{i = 1}^{N}\left[\mathlarger{\mathbbm{1}}_{i}^{\text{s}} \log (p_{i}^{\text{s}}) + (1-\mathlarger{\mathbbm{1}}_{i}^{\text{s}}) \log (1-p_{i}^{\text{s}})\right],  \nonumber
\\
& \:\:\:\:\:\:\:\:\:\:\:\:\:\:\:\:\text{s} = \text{start, end, entityness}
\\
\textbf{L}_{\text{type}} &=  -\frac{1}{N}\sum_{i = 1}^{N}\mathlarger{\mathbbm{1}}_{i}^{\text{entity}}\left[\sum_{c \in \textrm{classes}}\mathlarger{\mathbbm{1}}_{i}^{\text{c}}\log (p_{i}^{\text{c}})\right]
\end{align}
\normalsize

In Equation 1, $\alpha$ and $\beta$ are hyperparameters that control the weights of the boundary loss and the total entityness loss, and we used 0.5 and 1.0 as default values. In Equation 2-3, $i$ iterates through from sample 1 to N where each sample can be seen as a tuple $(sentence, index_{start}, index_{end})$. $\mathlarger{\mathbbm{1}}_{i}^{\text{entity}}$ is an indicator that the sample $i$ is an entity. $\mathlarger{\mathbbm{1}}_{i}^{\text{c}}$ indicates if the entity belongs to type c. Same idea for $\mathlarger{\mathbbm{1}}_{i}^{\text{s}}$. The start loss, end loss and entityness loss are all cross entropy loss over two classes. The type classification loss is only applied when the region proposal matches a true entity, and would be ignored otherwise.

The model structure is illustrated in Figure \ref{fig:mainModel}. For the start and end loss, we used a model that calculate multi-head dot products between the two tokens across the boundary. The intuition is that it needs to catch the relation signal between the two sides. After dot products we put an extra fully connected layer to transform the feature vector to the logits of a two-class classification. All heads have independent weights. 

For the entityness and type classification loss, we used the same architecture with separate weights, that is a fully connected layer followed by a max pooling layer, and then another fully connected layer to get the final logits. And we use ReLU activation function after each linear layer. The same as region proposal, we only consider the starting token in each word and mask out all trailing tokens within that word.

During inference, we look at the entityness probability first. Only if it's above a specified threshold will we look at the classification results and output the most likely type. 

\section{Related Work}
\label{section:related}
Named entity recognition (NER) is a classic problem in NLP. The goal of NER is to extract named entities from free text and these entities can be classified into several categories, for example person (PER), location (LOC) and geo-political entity (GPE). Traditionally NER is tackled by sequence labeling method. Many different models are developed along this direction, such as CRFs in \cite{lafferty2001conditional, sutton2007dynamic}, LSTM in ~\cite{hammerton2003named}, LSTM-CRF in ~\cite{lample2016neural} etc. More recently, people start using large-scale language models, such as BERT~\cite{devlin2018bert} and ELMo~\cite{elmo2018}. 

Nested named entity recognition takes the overlapping between entities into consideration~\cite{kim2003genia}. This cannot be easily done by traditional sequence labeling in that one can only assign one tag for each token. There have been many different approaches proposed for this problem. One important branch is the region-based method and our work can also be classified into this category.  ~\cite{finkel2009nested} leveraged parsing trees to extract subsequences as candidates of entities. ~\cite{xu2017fofe,sohrab-miwa-2018-deep} considers all subsequences of a sentence as candidates. A more recent paper by ~\cite{hongyu2019nugget} developed a model that locates anchor word first and then searches for boundaries of the entity, with the assumption that nested entities have different anchor words. Another work that is very close to ours is done by~\cite{zheng2019boundary}. In their paper they also proposed a two stage method. Our work is different than theirs from several perspectives. We have an entityness prediction in the second stage like the objectness in object detection, and thus we don't completely depend on the first stage to determine the region. And our model is built on BERT language model and finetunes all lower layers, while theirs is using LSTM plus pretrained word embedding. Another branch of researches is trying to design more expressive tagging schemas, some representative works are done by ~\cite{lu2015joint, kati2018nested, luwei2018overlap}. The current state of the art is ~\cite{li2019unified}, where they viewed the NER problem as a question answering problem and naturally solved the nested issue. Their model showed impressive power in both flat and nested NER tests. A major difference between our model and theirs is that they predict a paring score for any start and end index pair, which makes the feature matrix and the computational complexity a big issue. We instead predict entityness only for very few candidates, and we don't need to duplicate training examples for multiple queries, thus our training process takes much less time.

The main contribution of our work is that we bring up the idea to use a high-recall and relatively low-precision first stage model to select regions and use a more complicated model to predict the entityness and classification at the same time, with a global view of all the tokens spanning the candidate entity. Besides that, our model is much simpler and more lightweight than other similar models designed for nested NER tasks, and both training and inference run as fast as the plain BERT sequence labeling model. Our core model architecture is nothing but a linear layer plus a max pooling layer, but gives pretty good performance, especially on the ACE2005 dataset.

\section{Flat NER Experiments}
\label{section:experiment}

For the flat NER experiment, we used the CoNLL2003~\cite{conll2003ner} and OntoNotes 5.0~\cite{ontonotes5}. CoNLL2003 is an
English dataset with four types of named entities, namely Location, Organization, Person and Miscellaneous. And OntoNotes 5.0 is an English dataset containing text from many sources and including 18 types of named entity,

\begin{table*}[!h]
	\small
	\begin{center}
		\begin{tabular}{lrrrr}
			& Dev Precision & Dev Recall & Test Precision & Test Recall \\
			\hline
			\textbf{CoNLL2003 Region Proposal} &  &  & &   \\
			\hline
			BERT base & 71.3 & 98.0 & 70.7 & 96.2    \\
			BERT large & 71.4 & 98.0 & 70.6 & 96.5  \\
			\hline
			\hline
			\textbf{OntoNotes 5.0 Region Proposal} &  &  &  &   \\
			\hline
			BERT base (weight 0.5:0.5) & 69.3 & 90.8 & 69.3 & 89.6  \\
			BERT base (weight 0.3:0.7) & 67.8 & 92.7 & 67.4 & 92.2 \\
			BERT base (weight 0.2:0.8) & 66.3 & 93.9 & 65.8 & 93.7 \\
			BERT base (weight 0.1:0.9) & 63.8 & 95.1 & 63.1 & 95.5 
		\end{tabular}
	\end{center}
	\caption{\small \textbf{Region Proposal Model Results.} Precision and recall numbers are region metrics regardless of entity type. A prediction is correct as long as the region predicted matches the start and end of the ground-truth entity. Region precision and recall are reported for both dev and test sets.}
	\label{tab:rpn}
\end{table*}

\begin{table*}[!h]
	\small
	\begin{center}
		\begin{tabular}{lrrrrrr}
			& Dev Precision & Dev Recall & Dev F1 & Test Precision & Test Recall  &  Test F1  \\
			\hline
			\textbf{CoNLL2003} &  &  &  &   &  &   \\
			\hline
			BERT base  & 95.1 & 95.1  & 95.1 & 91.9 & \textbf{91.7} & \textbf{91.8}  \\
			BERT large & 96.1 & 95.4 & 95.8 & \textbf{92.0} & 91.6 & \textbf{91.8}  \\
			\hline
			\hline
			\textbf{OntoNotes 5.0} &  &  &  &  &  &   \\
			\hline
			BERT base & 87.7  & 87.3 & 87.5 & \textbf{86.9} & 86.5 & 86.7  \\
			BERT large & 88.0 & 88.3 & 88.2 & 86.8 & \textbf{87.3} & \textbf{87.0} 
		\end{tabular}
	\end{center}
	\caption{\small \textbf{Flat NER Results.} Standard NER precision and recall are reported here for both Dev and Test sets. We only showed the best model for each combination of dataset and the size of BERT encoder.}
	\label{tab:e2e}
\end{table*}

\subsection{First-Stage Training}
For region prediction, we used the default training parameters provided by HuggingFace Transformers~\cite{Wolf2019HuggingFacesTS} for token classification task, i.e. AdamW optimizer with learning rate=$5\times10^{-5}$, $\beta_{1}$=$0.9$ and $\beta_{2}$=$0.999$, hidden state dropout probability 0.1 etc. We finetuned both BERT-base-cased and BERT-large-cased models for 3 epochs with batch size of 64. The regions with length equal or less than 6 were selected as candidates for the second stage. We chose the threshold 6 because most named entites are shorter than 6 (99.9$\%$ in CoNLL2003 and 99.2$\%$ in OntoNotes 5.0). Since we ignore entity type in the first stage model, the precision and recall are based only on region proposals regardless of type. We keep the model to be as simple as possible because the only goal is to get high recall in the first stage. For the OntoNotes 5.0 model, a default training gave a model with pretty low recall, only 89.6$\%$, so we changed the weights in the cross entropy loss to raise the recall with a little trade-off of precision. Therefore precision and recall were reported at different weights for OntoNotes. The training results can be found in Table~\ref{tab:rpn}.

From the result, we can see that for CoNLL2003~\cite{conll2003ner}, base and large models gave pretty close p/r numbers, so we used BERT base in the following experiments. For OntoNotes dataset, we tried several different weights, it turned out that weights 0.4:0.6 and 0.3:0.7 both gave pretty good end-to-end results.

\begin{table*}[t]
	\small
	\begin{center}
		\begin{tabular}{lrrrrrr}
			&&\textbf{Genia}&&&\textbf{ACE2005}&\\
			\hline
			\textbf{Model} & \textbf{Precision($\%$)} & \textbf{Recall($\%$)} & \textbf{F1($\%$)} & \textbf{Precision($\%$)} & \textbf{Recall($\%$)} & \textbf{ F1($\%$)}\\
			\hline
			ARN~\cite{hongyu2019nugget} & 75.8 & 73.9 & 74.8 & 76.2 & 73.6 & 74.9 \\
			Boundary-aware Neural~\cite{zheng2019boundary} & 75.8 & 73.6 & 74.7 & - & - & - \\
			Merge-BERT~\cite{joseph2019merge} & - & - & - & 82.7 & 82.1 & 82.4  \\
			Seq2seq-BERT~\cite{strakova2019nested} & 80.1 & 76.6 & 78.3 & 83.5 & 85.2 & 84.3 \\
			Path-BERT~\cite{shibuya2019second} & 77.81 & 76.94 & 77.36 & 83.83 & 84.87 & 84.34 \\
			BERT-MRC~\cite{li2019unified} & 85.18 & 81.12 & \textbf{83.75} & 87.16 & 86.59 & \textbf{86.88} \\
			\hline
			\textbf{Our Model} &&&&&&\\
			\hline
			BERT Base Model & 76.6 & 75.1 & 75.9 & 82.8 & 84.9 & 83.8 \\
			BERT Large Model & 77.4 & 76.3 & 76.8 & 85.2 & 85.9 & 85.6 \\
			\hline
		\end{tabular}
	\end{center}
	\caption{\small \textbf{Nested NER Results on Genia and ACE2005.} }
	\label{tab:nested}
\end{table*}

\subsection{Second-Stage Training}

There are quite a few differences between our second-stage training and the traditional sequence labeling NER training, in the sense that our training is on region proposal level while sequence labeling is on sentence level. Each training example is now a combination of a sentence, a proposal start index and a proposal end index, and one sentence could emit multiple training examples. And the labels are no longer token labels, but labels designed for our specific loss, specifying if the start index is correct, if the end index is correct, if the region is corresponding to a true entity, and at last what the entity type is if all the previous answers are positive. The training samples are those region proposals output from the first model and we added more randomly selected negative regions to make it more robust, which turned out to be very important and will be explained with more details in the following paragraphs. Finally we evaluated the model performance in an end-to-end manner. Precision, recall and F1 score on CoNLL2003 and OntoNotes 5.0 can be found in Table \ref{tab:e2e}.

For the CoNLL2003 dataset~\cite{conll2003ner}, the best F1 score we got with base and large models are both 91.8, which is comparable but a little lower than the reported sequence labeling BERT results (BERT-Tagger). We didn't spend too much time on finetuning hyperparameters, it's possible that there is still room for the model. Another possible reason could be in our model we only predict for one entity at a time and didn't consider the interaction between entities when there are multiple in one sentence. For the OntoNotes 5.0 dataset we got F1 score 87.0. 

We tried a few tricks to improve the p/r number. The most effective one is to add more randomly selected regions as negative samples during the second stage. For each sentence, we generate one random region that has length in range [1, 6] and that doesn't fall into the existing candidates or true entities. We then labeled them as wrong samples and fed into the training process together with other candidates. With more random negative samples, the model is more robust when there is a big gap between the train and test set distributions, especially when we have a very strong stage-one model with high precision, which could have a strong bias on the distribution of negative samples. By adding more negatives, we have almost 0.7$\%$ gain in the F1 score, which will be shown with more details in the following ablation study section \ref{section:ablation}.

We also tried adding an extra loss to take into account the classification error for those regions that were not predicted completely correct but have a large overlap with one of the true entities. They were not exactly matching the ground true region, therefore we assigned a weight less than 1 (0.2 in our experiment). Intuitively, this solution is equivalent to adding more classification training data. However we didn't see significant improvement after this change. We also explored changing hyperparameters, for example the numbers of channels in the second-stage model. In the default settings we set the number of heads in dot product to be 32, and the dimension of feature vector in both entityness and classification to be 64. We tried half or double these numbers but the model performance degraded in both cases. More details can be found in the next ablation study subsection \ref{section:ablation}.

\section{Nested NER Experiments}
\label{section:nestedexp}
For nested NER experiments, we chose two mainstream datasets, the ACE2005 dataset~\cite{ace2005ner} and Genia dataset~\cite{kim2003genia}. The ACE2005 dataset contains 7 entity categories. For comparison, we follow the data preprocessing procedure in ~\cite{lu2015joint, luwei2018overlap, kati2018nested} by keeping files from bw, bn, nw and wl, and splitting these files randomly into train, dev and test sets by 8:1:1, respectively. For the Genia dataset, again we use the same preprocessing as ~\cite{finkel2009nested, lu2015joint} and we also consider 5 entity types - DNA, RNA, protein, cell line
and cell type.

Our training process for nested NER is basically the same as the previous section, since our model doesn't differentiate between flat and nested and will simply treat all entity regions in the same way even if they are overlapped. Considering the different distribution of entity lengths, we increased the length limit for region candidates from 6 words to 12 for ACE2005 and 8 for Genia. For the region proposal model, we again trained on top of BERT-base-cased for 3 epochs, and in the cross entropy loss we used weights 0.1:0.9. For the second stage, we trained both BERT-base and BERT-large for 10 epochs, the results are reported in Table~\ref{tab:nested}. With the BERT large model we got an average F1 score of 85.6 on ACE2005 and 76.8 on Genia. This is not as good as the current state-of-the-art result, but is quite competitive, and especially on ACE2005 dataset it's better than all other models except the BERT-MRC model.

\begin{table}[h]
	\small
	\begin{center}
		\begin{tabular}{|c|c|c|}
			\hline
			\textbf{Entity Type} & \textbf{Test Precision} & \textbf{Test Recall} \\
			\hline
			DNA & 74.8 & 72.6   \\
			\hline
			RNA & 90.6 & 82.1    \\
			\hline
			cell line & 78.5 & 67.1  \\
			\hline
			cell type & 73.6 & 72.2 \\
			\hline
			protein & 78.5 & 79.7  \\
			\hline
			Overall & 77.4  & 76.3  \\
			\hline
		\end{tabular}
	\end{center}
	\caption{\small \textbf{Precision and Recall by Category. (Genia)} }
	\label{tab:geniatype}
\end{table}

\begin{table}[h]
	\small
	\begin{center}
		\begin{tabular}{|c|c|c|}
			\hline
			\textbf{Entity Type} & \textbf{Test Precision} & \textbf{Test Recall} \\
			\hline
			PER & 88.6 & 89.9   \\
			\hline
			LOC & 77.0 & 76.2    \\
			\hline
			ORG & 74.8 & 77.8  \\
			\hline
			GPE & 87.7 & 86.7 \\
			\hline
			FAC & 82.0 & 77.8  \\
			\hline
			VEH & 72.3 & 71.4  \\
			\hline
			WEA & 75.0 & 73.8 \\
			\hline
			Overall & 85.2  & 85.9  \\
			\hline
		\end{tabular}
	\end{center}
	\caption{\small \textbf{Precision and Recall by Category. (ACE2005)} }
	\label{tab:acetype}
\end{table}

A detailed P/R analysis by category for Genia and ACE2005 datasets are given in Table~\ref{tab:geniatype} and Table~\ref{tab:acetype}.

\section{Ablation Study}
\label{section:ablation}
In this part, we did an ablation study to assess the contribution from each component of the new model. In the first experiment, we removed the start/end logits from the concatenated vector for entityness prediction and also removed the start/end loss from the total loss, to see if they are helpful to resolve boundary errors. Then, we evaluated the effects of the max pooling layer over the whole entity and used only the first token's feature vector instead. In the following experiments we also tried removing random negative samples and reducing the model size by using less channels or dimensions for the start/end unit, entityness prediction and entity classification etc. The evaluation results on the CoNLL2003 test set is shown in Table \ref{tab:ablation}.

From the results, we can see that the start/end prediction only has limited effect for the final F1, but with a more closer look at the errors we found that with start/end prediction, the boundary errors dropped from 149 (out of 5711 test samples) to 127 while the type classification errors changed from 222 to 230. The start/end loss did change the pattern of errors and corrected some boundary mistakes. Removing the max pooling layer brought an F1 drop greater than 1.0$\%$ and removing random negative samples brought 0.7$\%$ drop. After reducing the number of channels to half of the default, we got a drop of 0.5$\%$ and when cutting further, keeping only 25$\%$ channels, the model performance degraded pretty fast to 65$\%$, indicating a strong underfit.

\section{Error Analysis}
\label{section:error}
\begin{figure}[h]
	\centering
	\includegraphics[width=1.05\linewidth]{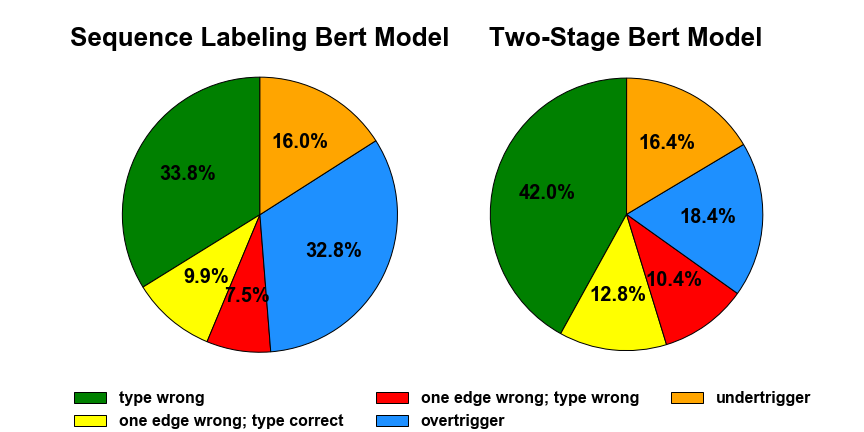}
	\caption{\small{\textbf{Error Analysis. } We have divided the entity recognition error to 5 classes, more details can be found in the text. We profiled the errors for traditional sequence labeling BERT model (the left panel) and also the two-stage model proposed in this paper (the right panel), to provide a deeper insight of the composition of model errors.}}
	\label{fig:error}
\end{figure}

To have a deeper understanding what type of errors our model is making, we did a further error profiling and also compared it with the result of a standard sequence labeling model based on BERT. Inspired by the methodology from Object Detection research, We divided the NER errors into the following types: (1) the region is correct but the type classification is wrong; (2) one of the left/right boundaries of the region is wrong, but classification is correct; (3)  one of the left/right boundaries of the region is wrong, and the classification is also wrong;  (4) over-triggering: the predicted entity has no overlap with any ground-truth entity; (5) under-triggering: for a true entity, either no overlapping region is proposed or the entityness model predicts no entity; (6) another type of error is that both boundaries are wrong but the region has overlap with at least one of true entities, in the evaluation we found that this type of error occurs only once, so we ignored it in the pie plot. We displayed the error composition for the new two-stage model as well as a traditional sequence labeling BERT model side by side, as shown in Figure \ref{fig:error}.

\begin{table}[t]
	\small
	\begin{center}
		\begin{tabular}{lrrr}
			& \textbf{Test P} & \textbf{Test R} & \textbf{F1}\\
			\hline
			\textbf{Original setting} & \textbf{91.9} & 91.7 & \textbf{91.8}    \\
			Remove start/end prediction & 90.4 & \textbf{91.9} & 91.7   \\
			Remove max pooling & 90.0 & 90.7 & 90.4 \\
			Remove random negatives & 90.6  & 91.6 & 91.1\\
			Remove 50$\%$ channels & 91.2 & 91.3 & 91.3 \\
			Remove 75$\%$ channels & 65.0 & 65.4 & 65.2\\
		\end{tabular}
	\end{center}
	\caption{\small \textbf{Ablation Study using CoNLL2003 Dataset.} All experiments are using the BERT base model and training with the same epochs. P, R and F1 are reported on test set.}
	\label{tab:ablation}
\end{table}

We ran analysis on two models both trained from BERT-base and having similar precision and recall. As we can see, for both models the dominant part is the type error (region is correct). We could consider using larger and more complicated type prediction models since there is a lot of room in that direction reading from the plot. The new model made significantly less over-triggering errors, which means the precision of the entityness prediction is good. And two models have similar amounts of single-boundary errors and under-triggering errors.

\section{Conclusion}

In this paper, we proposed a new two-stage model for named entity recognition. And many of the ideas are coming from the inspiration of object detection in computer vision. More specifically, with a coarse first-stage model to provide region proposals, we rely on a second-stage model to predict entityness and entity types at the same time. Through sufficient experiments in both flat and nested NER tasks, we found that it has better performance on nested NER, and we got F1 85.6 on ACE2005 dataset and F1 76.8 on Genia datase, better than many more complicated models. On flat NER tasks, it's still a few points behind the current SOTA results. In the future we are going to improve the model further to see where is the real limit of two-stage region-based named entity recognition.

\bibliography{ner_acl}
\bibliographystyle{acl_natbib}

\end{document}